\documentclass{article}
\usepackage{iclr2027_conference,times}
\usepackage{graphicx}
\usepackage{float}
\usepackage{booktabs,tabularx,array}
\usepackage{amsmath,amssymb,mathtools}
\usepackage{amsmath,amsfonts,bm}

\def\eqref#1{equation~\ref{#1}}
\def\1{\bm{1}}

\DeclareMathAlphabet{\mathsfit}{\encodingdefault}{\sfdefault}{m}{sl}
\SetMathAlphabet{\mathsfit}{bold}{\encodingdefault}{\sfdefault}{bx}{n}

\usepackage{hyperref}
\usepackage{url}

\newcommand{\model}{\textsc{GLaM}}
\newcommand{\systemname}{\textsc{GLaM Nav}}

\title{GLaM: Training a Latent World Model over Global Spatiotemporal Memory for Active Exploration and Navigation}

\author{
{\bfseries I-Tak Ieong$^{1,2}$, Ruizhi Feng$^{2}$, Zhaoyang Lu$^{2,3}$, Yifei Cao$^{2,4}$},Jiayao Zhao$^{2,5}$, Leon Li$^{\dagger,2}$,\\
{\bfseries  Senhua Zhu$^{2}$, Wenbo Ding$^{*,1}$}\\
\normalfont
$^{1}$ Tsinghua University\\
$^{2}$ Lab for Brain-Inspired Embodied Intelligence, EBKernel Technologies Co., Ltd\\
$^{3}$ Northeastern University\\
$^{4}$ University of California, Los Angeles\\
$^{5}$ Shanghai Jiao Tong University\\
$\dagger$ Project Leader \quad $^{*}$ Corresponding author\\
\texttt{Correspondence to: yangyd26@mails.tsinghua.edu.cn}
}

\iclrfinalcopy

\begin{document}
\maketitle

\begin{figure}[H]
\centering
\includegraphics[width=0.92\linewidth]{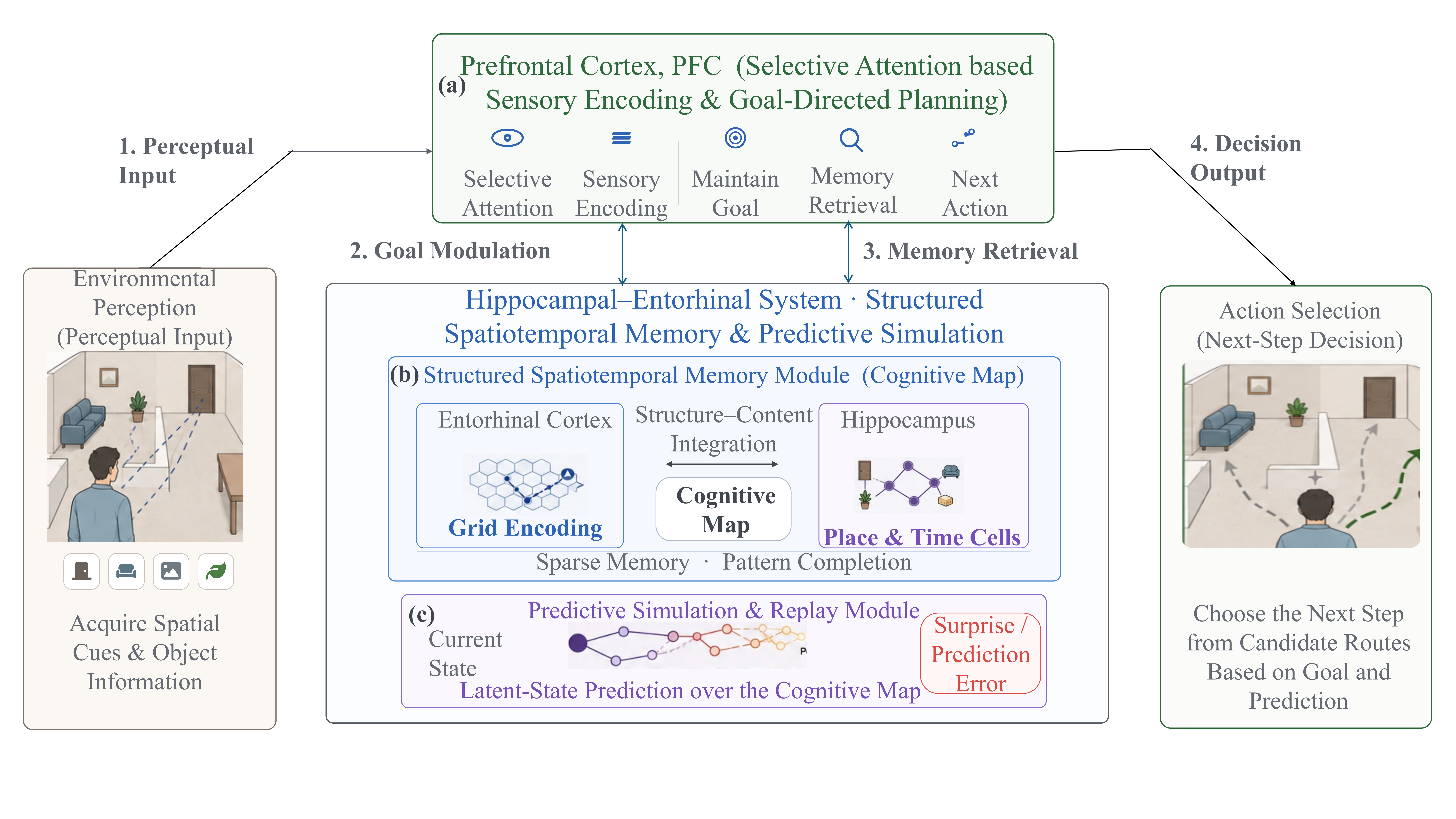}
\caption{Neurocognitive motivation for active exploration and navigation. The schematic summarizes the functional analogy underlying the proposed formulation: selective attention and goal maintenance guide perceptual encoding, hippocampal--entorhinal memory supports structured spatial representation and predictive simulation, and planning operates over retrieved working memory to select subsequent actions.}
\label{fig:motivation}
\end{figure}

\begin{abstract}
Active exploration and semantic navigation require an embodied agent to build memory from partial observations, predict how the evolution of observed spatial memory may support future motion, and convert that prediction into actionable plans. We present \model{}, a goal-conditioned latent world model trained over global spatiotemporal memory, and \systemname{}, the complete navigation system built around it. Given historical map tokens, a navigation goal, and the current robot pose, \model{} jointly predicts future map representations and robot-centric waypoint latents, allowing future spatial context and navigation intent to be inferred in a shared representation space. The model follows a JEPA-like latent prediction paradigm, operates directly on map-level latent tokens rather than RGB reconstruction, and uses a pretrained waypoint encoder--decoder to supervise and decode navigation plans within \systemname{}. Training data are collected by replaying ObjectNav expert trajectories in Habitat over HM3D v0.2 scene assets and slicing them into multi-timescale prediction samples. On a controlled HM3D-ObjectNav subset reproduction setting, \systemname{} improves over a reproduced BSC-Nav baseline in both success rate and success weighted by path length.
\end{abstract}

\section{Introduction}

Object-goal navigation requires an embodied agent to find an instance of a specified object category in an unfamiliar environment from partial egocentric observations. The target may initially be unobserved or leave the field of view as the agent moves, making navigation depend on both accumulated spatial knowledge and decisions about where to explore next. Semantic exploration methods demonstrate the value of explicit maps, object-related priors, and map-grounded subgoal selection~\citep{chaplot2020semexp,ramakrishnan2022poni,yokoyama2023vlfm}. These advances motivate a predictive use of memory: beyond retrieving what has already been observed, an agent should anticipate the evolution of observed spatial memory associated with further goal-directed motion.

Latent world models offer a representation-level approach to this problem. Joint-embedding prediction and pretrained-feature dynamics show that future states can be modeled without making pixel reconstruction the primary learning objective~\citep{assran2023self,zhou2025dinowm}. More broadly, predictive spatial representations have long been linked to flexible navigation and planning in cognitive-map accounts~\citep{stachenfeld2017hippocampus,pfeiffer2013hippocampal,behrens2018cognitive}. These developments motivate a concrete question for object-goal navigation: \emph{how can future spatial-memory prediction and goal-directed waypoint planning be learned jointly from a persistent, spatially indexed map?}

We present \model{}, a goal-conditioned latent world model that jointly predicts future map representations and robot-centric waypoint latents from global spatiotemporal memory. Here, global memory refers to an online representation of the explored environment rather than a complete or pre-built scene map. Given historical map tokens, a navigation goal, and the current robot pose, a shared backbone processes visual and waypoint queries in a single forward pass. The map branch predicts future visual-semantic map features, while the waypoint branch predicts a latent navigation plan. This formulation places spatial prediction and waypoint supervision in a common learning framework without requiring future RGB reconstruction.

The design is functionally motivated by cognitive-map accounts that connect structured spatial memory with prospective navigation~\citep{stachenfeld2017hippocampus,pfeiffer2013hippocampal}. As illustrated in Figure~\ref{fig:motivation}, we use persistent spatial representation and memory-based prediction as computational design principles rather than claiming a direct implementation of biological neural circuits. At the system level, we integrate \model{} into \systemname{}, which combines online mapping, goal-conditioned memory retrieval, observation-based target verification, and geometric motion control. Predicted representations provide prospective context, whereas real observations remain the source of memory updates and target verification.

Training uses expert-controlled ObjectNav trajectories collected in Habitat over HM3D scene assets~\citep{savva2019habitat,yadav2022hm3dsem}. We construct temporally aligned samples containing historical map tokens, future map targets, and expert waypoints at multiple prediction intervals. A waypoint encoder--decoder is pretrained through reconstruction and then frozen, providing a latent supervision space and a corresponding decoding interface. The world model is trained with masked regression objectives for future-map features and waypoint latents. Its predictions therefore concern future observations and plans represented in the expert data, rather than unrestricted counterfactual outcomes for arbitrary action sequences.

We evaluate the resulting navigation system against a reproduced BSC-Nav baseline~\citep{ruan2026brain} on the same HM3D-ObjectNav evaluation subset. As reported in Table~\ref{tab:hm3d_results}, \systemname{} increases success rate from 78.50\% to 86.89\% and SPL from 47.70\% to 48.35\%. We also provide qualitative real-world demonstrations of online exploration, goal navigation, and reuse of previously acquired spatial memory. These experiments assess the navigation performance of the integrated system and the feasibility of reusing its spatial memory with real sensing and motion.

Our contributions are threefold: (i) a map-grounded latent model that jointly predicts future spatial-memory features and goal-conditioned waypoint latents; (ii) a training formulation combining multi-timescale trajectory supervision with a pretrained waypoint-latent interface; and (iii) an integrated navigation system evaluated through a matched-episode simulation comparison and qualitative real-world demonstrations.

\section{Related Work}

Goal-directed semantic exploration methods such as SemExp \citep{chaplot2020semexp}, PONI \citep{ramakrishnan2022poni}, and VLFM \citep{yokoyama2023vlfm} show that explicit maps and exploration policies are effective for object-goal navigation. Earlier navigation systems and planners, including cognitive mapping and planning, semi-parametric topological memory, target-driven visual navigation, distributed visual navigation policies, vision-and-language navigation, and Active Neural SLAM, established the value of explicit memory, long-horizon subgoal selection, semantically grounded planning, and map-grounded control \citep{gupta2017cmp,savinov2018semi,zhu2017targetdriven,wijmans2020ddppo,anderson2018vln,chaplot2020learning,kuipers2000spatial}. These methods motivate the use of map structure and semantic cues, but they do not explicitly learn future map representations and waypoint plans in a shared latent predictive model.

Structured spatial memory is also central to recent navigation systems. ConceptGraphs \citep{gu2023conceptgraphs} organizes open-vocabulary scene content into a persistent 3D scene graph, while BSC-Nav \citep{ruan2026brain} builds a brain-inspired spatial intelligence pipeline around landmark and survey memories. These directions are supported by large-scale embodied-AI simulators and 3D scene resources such as Habitat, HM3D-Semantics, and Matterport3D, which make persistent map construction and evaluation practical at scale \citep{savva2019habitat,yadav2022hm3dsem,yadav2023habitat,chang2017matterport3d}. They motivate persistent map structure and retrieval, but they do not directly cast future map evolution and waypoint planning as a joint latent prediction problem over global spatiotemporal memory.

Latent world models provide the predictive side of our design. I-JEPA \citep{assran2023self} predicts informative target representations rather than pixels, while DINO, MAE, and DINOv2 illustrate the strength of large-scale self-supervised visual features as a substrate for downstream representation learning \citep{caron2021dino,he2022mae,oquab2024dinov2}. DINO-WM \citep{zhou2025dinowm}, PlaNet \citep{hafner2019planet}, and MuZero \citep{schrittwieser2020muzero} show how learned latent dynamics can support planning without reconstructing raw observations at every stage, and Banino et al.\ \citep{banino2018vector} further show that structured spatial codes can emerge in agents trained for navigation. Our focus is to bring this predictive perspective into semantic navigation by jointly modeling future map tokens and waypoint latents from shared global-memory context.

\section{Active Exploration and Navigation System Based on \systemname}
\label{sec:system}

\begin{figure}[t]
\centering
\includegraphics[width=\linewidth]{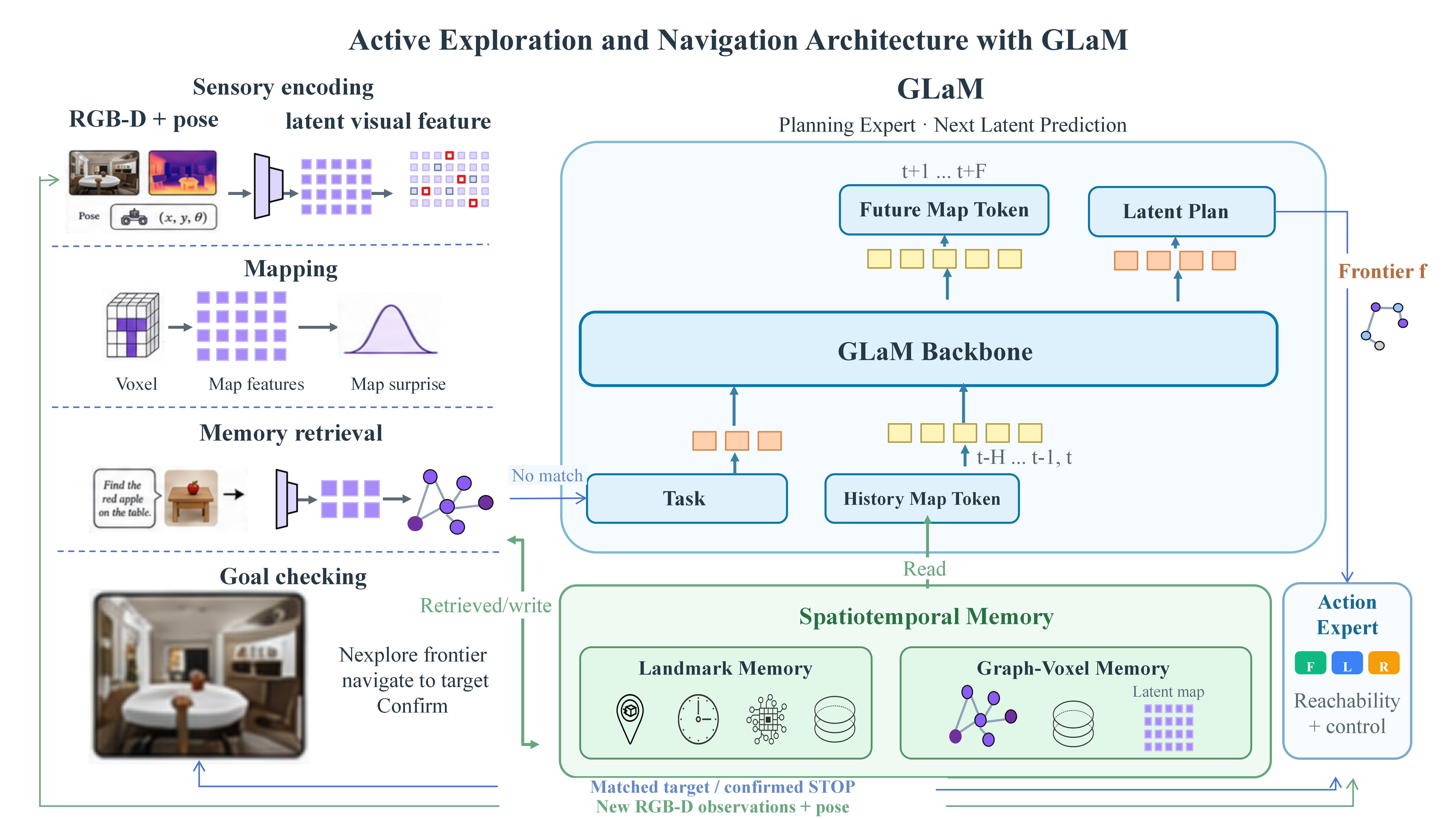}
\caption{Overview of the \systemname{} architecture for active exploration and object-goal navigation. The system closes the loop among sensory encoding, mapping, memory retrieval, structured spatiotemporal memory, \model{}-based latent prediction, and action execution. Real RGB-D observations update memory, while the learned predictive module supplies future map tokens and latent plans that guide navigation decisions.}
\label{fig:glam_arch}
\end{figure}

Figure~\ref{fig:glam_arch} illustrates the overall organization of \systemname{}. At the system level, \systemname{} integrates four interacting parts: perception and retrieval, structured spatiotemporal memory, the \model{} predictive module, and a planning-and-control layer that grounds latent predictions into executable motion. Throughout the paper, \model{} refers to the learned latent world model, whereas \systemname{} refers to the complete navigation system built around that model.

The \textbf{Perception, Retrieval, and Goal Verification} module is responsible for perceptual encoding, working-memory querying, and target verification. It receives RGB-D observations, the agent pose, and the motion state, and converts them into semantically useful latent features. These features are projected into the global map structure used by the rest of the system. The same module also retrieves goal-relevant map context and decides whether the target has already been verified in current observation or memory. If verified, the system navigates directly toward the retrieved location. Otherwise, it triggers predictive exploration and planning.

The \textbf{Spatiotemporal Memory Module} is responsible for memory construction, storage, updating, and retrieval. Selected latent visual features, depth-projected coordinates, instance information, and robot poses are organized into a sparse spatial memory that combines landmark memory with graph-voxel memory. During memory updating, the system compares new observations with existing memory and decides whether the corresponding units should be added, fused, or refreshed. During retrieval, semantic information together with its associated latent features is used to identify the most relevant candidate navigation locations from the current memory.

The \textbf{\model{} backbone} is the predictive core of \systemname{}. It takes historical map tokens, the navigation goal, and the current robot pose as input and predicts future map representations together with latent waypoint plans. The model does not reconstruct future RGB frames. Instead, following the latent-space learning logic of the original Cog-WM design, it predicts high-level latent map representations and uses those predictions to support planning. In this formulation, \model{} jointly learns two outputs: future spatial-memory prediction and goal-conditioned waypoint-latent prediction. The first output estimates how global memory is expected to evolve over future horizons. The second output represents navigation plans in a robot-centric latent space that can later be decoded into waypoint increments.

The \textbf{Planning Expert} receives retrieved map context and the future map and waypoint predictions produced by \model{}. It proposes or grounds a feasible subgoal in the current observed map and sends executable commands to the low-level controller. This preserves a clear boundary between latent prediction and physical control: \model{} predicts future map structure and navigation intent, while the rest of \systemname{} remains responsible for geometric execution, reachability checking, and collision-aware motion.

The full navigation loop is as follows. \systemname{} first surveys its surroundings and updates the online map. It then retrieves relevant regions from landmark memory and graph-voxel memory based on the goal. If the goal has already been verified, the system navigates directly to that location. Otherwise, it uses \model{} to predict future map context and waypoint latents from the current memory state, decodes the waypoint plan, and executes the selected motion branch. After execution, the system updates memory with real observations only, verifies the goal again, and repeats the exploration--retrieval cycle until success or termination.

\section{\model{}: Training a Latent World Model over Global Spatiotemporal Memory}
\label{sec:glam_model}

As shown in Figure~\ref{fig:glam_arch}, \model{} should be understood as the predictive core inside \systemname{}. Within the full navigation loop, \model{} receives task condition and historical map tokens, interacts with the spatiotemporal memory modules, and produces two coupled outputs: future map-token prediction and latent planning representations. These outputs provide predictive evidence before action grounding in \systemname{}, while memory construction, retrieval, and control remain system-level functions.

\subsection{Model Training Paradigm}
\label{sec:glam_paradigm}

GLaM learns a goal-conditioned relationship between global spatiotemporal memory, future map states, and subsequent navigation waypoints. Its functional organization is inspired by spatial indexing, place--event binding, and predictive replay in the hippocampal--entorhinal system, together with goal-conditioned memory retrieval and planning control associated with the prefrontal cortex~\citep{whittington2020tolman,pfeiffer2013hippocampal,eichenbaum2017prefrontal}. These correspondences motivate the functional organization of the model rather than a direct implementation of biological neural circuits.

Given historical map tokens, a goal description, and the current robot pose, GLaM jointly predicts future map representations and robot-centric waypoint latents over a fixed-length horizon:
\begin{equation}
\left(
\widehat{\boldsymbol Z}_{t+1:t+F},
\widehat{\boldsymbol E}^{w}_{t+1:t+F}
\right)
=
F_{\Theta}\!\left(
\boldsymbol Z_{t-H+1:t},
 g,p_t,
\boldsymbol Q^{v},\boldsymbol Q^{w}
\right).
\label{eq:glam_joint_prediction}
\end{equation}
Here, $\boldsymbol Z_{t-H+1:t}$ denotes the historical map-token sequence, $g$ is the goal description, and $p_t$ is the current robot pose. The parameters $H$ and $F$ specify the history length and prediction horizon, respectively. The learnable queries $\boldsymbol Q^{v}$ and $\boldsymbol Q^{w}$ correspond to future-map prediction and waypoint-latent prediction. The two outputs, $\widehat{\boldsymbol Z}_{t+1:t+F}$ and $\widehat{\boldsymbol E}^{w}_{t+1:t+F}$, represent the predicted future map tokens and waypoint latents, respectively.

GLaM follows a JEPA-like latent-space prediction paradigm~\citep{assran2023self}: it directly regresses continuous future representations rather than reconstructing future RGB observations. Visual and waypoint queries are supplied together in a single forward pass, allowing both outputs to be predicted from the same memory, goal, and pose context. The predicted waypoint latents are subsequently decoded into normalized planar pose increments, as described in Section~\ref{sec:glam_encoding}. We evaluate the practical utility of this joint predictive representation through the downstream tasks of active exploration and semantic navigation.

\subsection{Training Datasets}
\label{sec:glam_training_data}

Training data are collected by replaying ObjectNav expert trajectories in Habitat using HM3D v0.2 scene assets. HM3D provides large-scale reconstructed indoor environments for simulated semantic-navigation interactions, including RGB-D observations, robot poses, and navigation meshes~\citep{yadav2023habitat}.

During collection, each episode preserves the original scene, start pose, goal category, and expert rollout defined by the benchmark. Along each expert trajectory, we record RGB-D observations, robot poses, executed actions, and the map features derived from these observations, while also storing the frontier sequence encountered during exploration together with the updated map features associated with frontier progression. We then slice the replayed trajectories into training pairs with historical map tokens, future map targets, and future waypoint latents. This keeps the description of data construction explicit while maintaining the same sample definition used by the model.

The collected subset contains 577 ObjectNav episodes across 145 scenes, comprising 55,392 recorded interaction steps. At an observation rate of 1 frame per second, these steps correspond to approximately 15.4 hours of observations. Slicing the trajectories with multiple prediction intervals and history windows produces 12,101 multi-timescale prediction samples. These statistics describe the subset collected for this study rather than the full HM3D dataset.

\subsection{Map-Token and Waypoint Latent Encoding}
\label{sec:glam_encoding}

\paragraph{Map-token encoding.}
Map tokens are constructed from four RGB-D views. Visual features are extracted from the observations, projected into a voxel grid, and fused within each voxel. Voxel coordinates are expressed in the current robot frame and encoded using three-dimensional spatial rotary positional embeddings (RoPE). The resulting token sequences are truncated or zero-padded to a fixed length, with validity masks identifying non-padded positions. A linear input projection maps the visual features to the 2560-dimensional representation used by the backbone.

\paragraph{Waypoint latent encoding.}
Each waypoint is represented as a robot-centric planar pose increment $(\Delta x,\Delta y,\Delta\theta)$. We normalize its translation and encode its heading using a sine--cosine representation:
\begin{equation}
\begin{aligned}
\boldsymbol u
&=
\left[
\operatorname{clip}\!\left(\frac{\Delta x}{s_x},-1,1\right),
\operatorname{clip}\!\left(\frac{\Delta y}{s_y},-1,1\right),
\sin\Delta\theta,
\cos\Delta\theta
\right]^{\!\top},\\
\boldsymbol e^{w}
&=\phi(\boldsymbol u)\in\mathbb R^{2560}.
\end{aligned}
\label{eq:glam_waypoint_encoding}
\end{equation}
The translation scales $s_x$ and $s_y$ are estimated exclusively from the training split. The encoder $\phi$ is a lightweight two-layer multilayer perceptron with dimensions $4\rightarrow d_a\rightarrow2560$, followed by LayerNorm, where $d_a$ denotes the hidden-layer width.

Before world-model training, $\phi$ is jointly pretrained with a lightweight decoder $\rho$ by minimizing the mean squared error between normalized waypoints $\boldsymbol u$ and their reconstructions $\rho(\phi(\boldsymbol u))$. This pretraining is independent of the world model. Both modules are subsequently frozen: the encoder provides waypoint-latent supervision during world-model training, and the decoder converts predicted latents into normalized waypoint representations at inference:
\begin{equation}
\widehat{\boldsymbol u}_j
=\rho\!\left(\widehat{\boldsymbol e}^{w}_j\right).
\label{eq:glam_waypoint_decoding}
\end{equation}
Inference uses the same decoder $\rho$ that was pretrained jointly with $\phi$.

\subsection{Training Loss Design}
\label{sec:glam_training_loss}

During world-model training, Qwen3.5-0.8B backbone, input projections, learnable queries, and two regression heads are optimized. The objective combines masked mean squared errors for future-map prediction and waypoint-latent prediction:
\begin{equation}
\mathcal L
=
\lambda_v\,
\operatorname{MSE}_{m^v}\!\left(
\widehat{\boldsymbol Z},\boldsymbol Z^*
\right)
+
\lambda_w\,
\operatorname{MSE}_{m^w}\!\left(
\widehat{\boldsymbol E}^{w},\phi(\boldsymbol U^*)
\right).
\label{eq:glam_training_objective}
\end{equation}
Here, $\boldsymbol Z^*=\{\boldsymbol z_i^*\}$ contains future-map targets produced by the frozen visual encoder and deterministic map-construction pipeline. The sequence $\boldsymbol U^*=\{\boldsymbol u_j^*\}$ contains normalized expert waypoints, with $\phi$ applied to each waypoint to obtain its target latent representation. The binary masks $m_i^v$ and $m_j^w$ identify valid visual and waypoint target positions, respectively, while $\lambda_v$ and $\lambda_w$ control the relative contributions of the two objectives.

For $d$-dimensional target vectors $\boldsymbol x_\ell^*$ and a binary validity mask $m$, masked MSE is defined as
\begin{equation}
\operatorname{MSE}_{m}\!\left(
\widehat{\boldsymbol X},\boldsymbol X^*
\right)
=
\frac{
\sum_{\ell}m_{\ell}
\left\|\widehat{\boldsymbol x}_{\ell}-\boldsymbol x_{\ell}^{*}\right\|_2^2
}{
d\sum_{\ell}m_{\ell}
},
\qquad \sum_{\ell}m_{\ell}>0.
\label{eq:glam_masked_mse}
\end{equation}
Thus, each objective is normalized by both its number of valid target positions and the dimensionality of its target representation. The waypoint-latent loss uses $d=2560$, whereas the map-prediction loss uses the dimensionality of the future-map target features. Both objectives supervise continuous representations directly; no language-model cross-entropy term is used.

\section{Experiments}
\label{sec:experiments}

\subsection{Experimental Setting}

We evaluate \model{} in simulation on the HM3D ObjectNav benchmark \citep{yadav2023habitat}, a standard embodied object-navigation benchmark in the Habitat ecosystem built on HM3D-Semantics v0.2 \citep{yadav2022hm3dsem}. The benchmark contains 216 semantically annotated real-world indoor 3D reconstructions, split into 145 training scenes, 36 validation scenes, and 35 test scenes. Scenes cover apartments, offices, and other multi-room indoor layouts, and target categories include six common indoor objects such as chair, sofa, potted plant, bed, toilet, and TV. Each episode specifies a scene, an initial robot pose, a target category, and corresponding target instances. During evaluation, the agent starts from a random pose and orientation and must complete exploration, target retrieval, and path planning from RGB-D observations without access to a pre-built map or privileged target coordinates.

Following the standard ObjectNav protocol, an episode is counted as successful when the agent executes the STOP action within the allowed budget, is within $1.0\,\mathrm m$ of a valid target instance, and the target is observable from the stopping location. To enable a controlled comparison against a strong brain-inspired baseline and to reduce reproduction overhead, we evaluate on a 50\% subset of the HM3D benchmark under the same split and protocol. We reproduce BSC-Nav \citep{ruan2026brain} as the reference method and then run \model{} on the same evaluation episodes, so that both methods are compared under matched scene distributions, target categories, and task difficulty. Under this subset reproduction setting, we do not directly compare the reported numbers with methods evaluated on the full benchmark, different HM3D versions, different input modalities, or different evaluation protocols.

We report Success Rate (SR) and Success weighted by Path Length (SPL). SR measures the fraction of episodes in which the agent successfully reaches the target and stops correctly within the allowed budget, and therefore primarily reflects whether the target can be found reliably. SPL additionally accounts for path efficiency, so it decreases when the agent succeeds but follows an unnecessarily long route. In addition to these primary metrics, final reports should include path length to first verified target observation, collision rate, timeout rate, false stop rate, and runtime including both the frozen feature service and the trainable predictor. For the training set itself, reporting should cover the number of scenes, replayed episodes, recorded interaction steps, and resulting multi-timescale prediction samples. Where additional learner-visited states are introduced in future data aggregation experiments, they should be reported separately rather than merged into the original expert-only manifest, following the logic of no-regret dataset aggregation in imitation learning \citep{ross2011dagger}.

\subsection{HM3D-ObjectNav Results}

\begin{table}[t]
\centering
\caption{HM3D-ObjectNav results on the 50\% subset reproduction setting. Relative improvements are computed with respect to the reproduced BSC-Nav baseline on the same evaluation episodes.}
\label{tab:hm3d_results}
\begin{tabular}{lcc}
\toprule
Method & SR (\%) & SPL (\%) \\
\midrule
BSC-Nav~\citealp{ruan2026brain} & 78.50 & 47.70 \\
\systemname{} (ours) & 86.89 & 48.35 \\
\midrule
Relative improvement & 10.69\% $\uparrow$ & 1.36\% $\uparrow$ \\
Absolute gain & +8.39 pts & +0.65 pts \\
\bottomrule
\end{tabular}
\end{table}

Table~\ref{tab:hm3d_results} reports higher SR and SPL for \systemname{} than for the reproduced BSC-Nav baseline on the matched evaluation subset. SR increases from 78.50\% to 86.89\%, a gain of 8.39 percentage points, while SPL increases from 47.70\% to 48.35\%, a gain of 0.65 percentage points. These point estimates indicate a larger improvement in task completion than in the aggregate path-efficiency metric. They do not, by themselves, establish shorter trajectories on commonly successful episodes or isolate the contribution of future-map prediction. Paired trajectory analysis and component ablations are required to assess those explanations.

\subsection{Qualitative Real-World Demonstrations}
\label{sec:real_world_deployment}

We further evaluate whether the exploration and navigation behavior of \systemname{} transfers from simulation to a physical indoor environment. The system was deployed in an open-plan research office on an Astribot S1 wheeled dual-arm humanoid equipped with an NVIDIA Jetson AGX Orin, an RGB-D camera, LiDAR, and proprioceptive sensors. Our implementation separates the body-independent global memory and latent prediction modules from the body-specific motion controller. The predictive model therefore communicates with the robot through a metric navigation interface, while the platform controller executes the resulting motion commands. The experiments reported here validate this interface on Astribot S1.

\begin{figure*}[t]
    \centering
    \IfFileExists{figs/fig_real_robot_combined_vertical.png}{%
        \includegraphics[height=0.76\textheight,keepaspectratio]{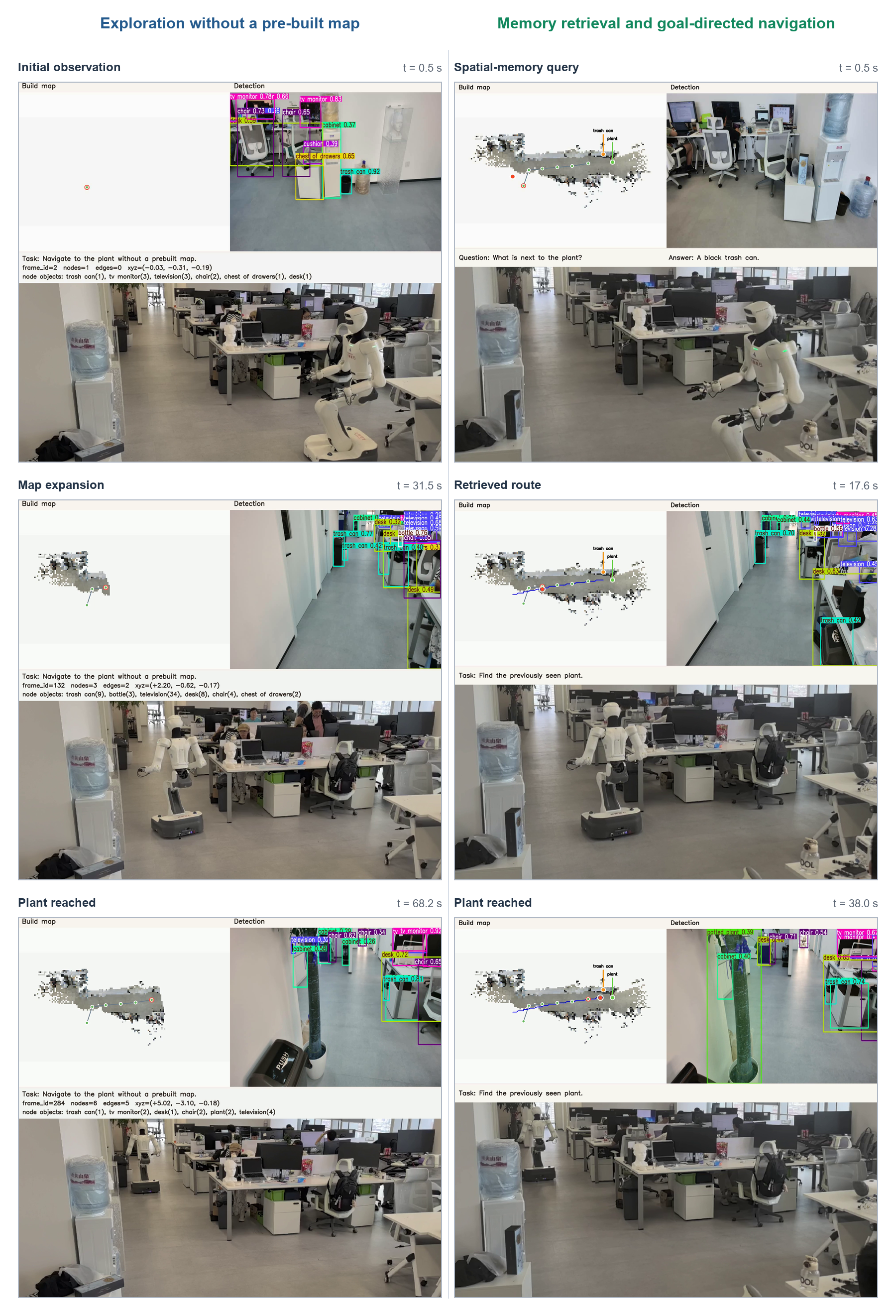}%
    }{%
        \fbox{\parbox[c][0.36\textheight][c]{0.68\textwidth}{\centering
        Upload \texttt{fig\_real\_robot\_combined\_vertical.png} to \texttt{figs/}.}}%
    }
    \caption{Real-world deployment of \systemname{} on an Astribot S1. Left: the robot explores without a pre-built map, expands its spatiotemporal memory, and reaches the plant region. Right: the system answers a spatial-memory query, retrieves the historical plant location, and navigates back to it. Each frame combines the online map, first-person semantic perception, and an external view of the robot.}
    \label{fig:real_robot}
\end{figure*}

\subsubsection{Active Exploration and Semantic Navigation}

The first experiment instructs the robot to navigate to a plant in an initially unknown environment. The robot receives no pre-built 2D or 3D map and no target coordinates. It incrementally associates visual observations with spatial nodes while moving through the office, producing a global spatiotemporal memory whose topology expands with the explored free space. As shown in the left column of Figure~\ref{fig:real_robot}, the map grows from a single initial node to a connected route containing newly observed semantic landmarks. The robot then reaches the plant region using the same online memory for localization, retrieval, and planning. This sequence demonstrates that \systemname{} can couple map construction with goal-directed motion in a real environment rather than relying on an environment-specific map prepared before deployment.

\subsubsection{Spatiotemporal Memory Retrieval and Spatial Question Answering}

The second experiment tests whether information acquired during exploration remains useful after the target leaves the current field of view. The robot is first asked what is located next to the plant. By retrieving the stored semantic and spatial relations, the system answers that a black trash can is beside the plant. The robot then receives the instruction to find the previously observed plant. As shown in the right column of Figure~\ref{fig:real_robot}, \systemname{} retrieves the corresponding historical location, converts it into a route over the accumulated memory, and navigates back to the target region. The result shows that the map stores more than an instantaneous detector output: it supports persistent object-location associations and local spatial relations that can be reused for both navigation and question answering.

Together, the two experiments provide complementary evidence for real-world operation. The first evaluates online memory construction under active motion, while the second evaluates delayed retrieval and reuse of previously acquired spatial knowledge. Although a broader quantitative study across buildings and robot embodiments remains necessary, these demonstrations show that the same memory and prediction interface used in simulation can operate with real sensing and robot motion without a pre-built map.

\subsection{Ablation Studies}

Detailed ablation studies and additional experimental results will be released in a subsequent public version of the manuscript and project materials.

\section{Limitations}

\paragraph{Prediction scope and data coverage.}
\model{} learns from expert-controlled trajectories, so its supervision covers the states and behaviors represented in those rollouts rather than all states that the deployed policy may encounter. Recovery from navigation errors, unfamiliar exploration patterns, and alternative routes may therefore require experience not represented in the current training set. Moreover, the model conditions on memory, a goal, and pose rather than an externally specified future action sequence. Its joint predictions should consequently be interpreted as goal-conditioned map and waypoint estimates, not as a general simulator of arbitrary action-conditioned futures.

\paragraph{Uncertainty and control feasibility.}
The current model produces deterministic latent predictions over a fixed horizon. Under partial observability, the same observed memory may admit multiple unseen layouts and feasible routes, but the regression objective does not explicitly represent these alternatives or calibrate predictive uncertainty. Furthermore, low map-feature or waypoint-latent error does not guarantee accurate target localization, collision-free motion, or consistency between the two predicted outputs. The system therefore relies on observation-based verification and geometric control rather than treating a latent prediction as a verified environmental fact.

\paragraph{Memory fidelity and capacity.}
Spatial memory depends on visual features, depth projection, and pose alignment. Errors in these components can affect both retrieval and the context supplied to the predictor. Although the stored map is persistent, the model accesses a bounded token representation; truncation may omit spatial detail or previously observed evidence as exploration expands. This creates a trade-off between memory coverage, token resolution, and inference cost. The present evaluation does not establish robustness to substantial localization drift, persistent scene changes, or substantially larger environments.

\section{Conclusion}

\model{} combines future-map prediction and goal-conditioned waypoint-latent prediction in a single latent world model over global spatiotemporal memory. By training on replayed HM3D ObjectNav trajectories and supervising both future map tokens and waypoint latents, the model connects memory, prediction, and planning in one representation space. The HM3D-ObjectNav subset evaluation suggests that this formulation improves target-search reliability while maintaining competitive path efficiency against a strong structured-memory baseline.

\bibliography{iclr2027_conference}
\bibliographystyle{iclr2027_conference}

\end{document}